\documentclass[10pt,twocolumn,letterpaper]{article}

\usepackage{wacv}
\usepackage{times}
\usepackage{epsfig}
\usepackage{graphicx}
\usepackage{amsmath}
\usepackage{amssymb}

\wacvfinalcopy 

\def\wacvPaperID{11} 

\ifwacvfinal\fi
\begin{document}

\title{Group Affect Prediction Using Multimodal Distributions}

\author{Saqib Nizam Shamsi \\
Aspiring Minds\\
{\tt\small shamsi.saqib@gmail.com}
\and
Bhanu Pratap Singh \\
Univeristy of Massachusetts, Amherst\\
{\tt\small bhanupratap.mnit@gmail.com}
\and
Manya Wadhwa \\
Johns Hopkins University\\
{\tt\small mwadhwa1@jhu.edu}
}

\maketitle
\ifwacvfinal\thispagestyle{empty}\fi

\begin{abstract}
We describe our approach towards building an efficient predictive model to detect emotions for a group of people in an image. We have proposed that training a Convolutional Neural Network (CNN) model on the emotion heatmaps extracted from the image, outperforms a CNN model trained entirely on the raw images. The comparison of the models have been done on a recently published dataset of Emotion Recognition in the Wild (EmotiW) challenge, 2017. The proposed method \footnote{Code for the approach and experiments is available at: \texttt{https://github.com/saqibns/cv-aal-2018}} achieved validation accuracy of 55.23\% which is 2.44\% above the baseline accuracy, provided by the EmotiW organizers.
\end{abstract}

\section{Introduction}

Emotion categorization of a face has always been an interesting problem and researchers have developed various approaches to solve this problem effectively. Categorizing the emotion of an image which contains multiple people has also gained attention \cite{emotiw_challenge}. The images are of real-life scenarios such as an image of a protest or of children playing in a park or some friends sitting and having a discussion. The problem becomes more difficult than emotion categorization of a facial image because one needs to address multiple sub-problems. For instance, the backgrounds of images might be starkly different from each other, the faces of the people may be occluded or they might not be facing the camera. Single-faced images are usually front-facing images in which faces are easier to crop out from the original image \cite{Goodfellow2013}. With more pictures being posted every day on social media sites such as Instagram and Facebook, this problem has interesting applications such as analyzing the emotion of a group of people attending an event which could prove helpful in tagging the pictures automatically.

In this paper, we discuss our approach for predicting the emotion of a group of people in an image. The goal is to build a model which could classify the emotion of a group of people in an image into one of the three classes: positive, negative and neutral. The dataset, also known as Group Affect Database 2.0 \cite{the_more_the_merrier}, contains images from different social events in all the three classes such as convocations, marriages, meetings and funerals. The dataset was a part of a challenge, which included different sub-problems from previous challenges. Like the 4\textsuperscript{th} EmotiW challenge, where the goal was to predict the happiness index of a group image, this challenge dealt with group images. It was also similar to the 3\textsuperscript{rd} EmotiW challenge, where the goal was to predict the emotion of a single face in an image. The difference was that there were seven categories of emotions in the 3\textsuperscript{rd} EmotiW challenge. In this challenge there was a group of people in an image and the emotion of the image was to be classified into one of the three classes: Positive, Netural and Negative.

Researchers have used many interesting approaches for this problem, such as in \cite{zhang2014representation}, the researchers had tried a dimensional approach where facial expressions were treated as a regression problem in the Arousal-Valence space. In \cite{DBLP_conf_icmi_NgNVW15}, a two-stage fine-tuning was applied on deep CNN while doing transfer learning. In \cite{Yu_2015_IBS}, multiple deep network training layers are utilized for emotion prediction for the 3\textsuperscript{rd} EmotiW challenge. It is interesting to see that all the three winners of the 3\textsuperscript{rd} EmotiW challenge utilized deep learning networks in their techniques \cite{Yu_2015_IBS}\cite{DBLP_conf_icmi_NgNVW15}\cite{kim2015hierarchical}. This indicated that deep networks could be helpful for the dataset of the 5\textsuperscript{th} challenge as well. The main difference was in the aggregation of emotions of multiple people from the image to predict the mood/expression of the image in general. 

In this paper, we use a bottom-up \cite{gibson_topdown_bottomup} approach for emotion detection. We perform face level emotion detection. We then experiment with average estimations (section 4.1.1) but as discussed in \cite{happiness_intensity}, which is also mentioned in \cite{DBLP_conf_icmi_VonikakisYN016}, averaging is not a reliable measure to predict group level emotion. We also train random forests (section 4.1.2) but the performance leaves much to be desired. Next, we use the face level detections and combine individual images by constructing heatmaps for emotion intensity. We then train Convolutional Neural Networks on the heatmaps (section 4.2). As mentioned in  \cite{the_more_the_merrier}, large size of faces with smiles play a role in determination of affect of a group. For example, a face closer to the camera, thus having a bigger size, would have more contribution towards the overall affect of the image as compared to the one that is farther away and is thus smaller. To that end, we create heatmaps which are normalized by the distance to the center of the image and use them to train CNNs (section 4.2.3). Finally, we demonstrate that training CNNs on heatmaps gives better results than training them on raw images (section 4.3).

\section{Dataset}
The group emotion recognition dataset \cite{emotiw_dataset} was provided as a part of 5\textsuperscript{th} EmotiW challenge \cite{emotiw_challenge}. The dataset was divided into three parts: train, validation and test. However, we use only the training and validation sets for the current work. This is because the test set is unlabeled and was to be used to judge submissions of the challenge. Since it was unlabeled we could not use it to judge the performance of our models. 

Each set consists of three type of group images: images with positive emotion such as of marriages and party, neutral images such as of meetings and images with negative emotion such as of protests and funerals. The distribution of images in each of the sets is presented in table ~\ref{table:dataset}.

For training the networks, we divide the training set into two parts out of which one is used for training the network and the other is used as a hold-out set for model selection. We use stratified sampling and pick 10\% of the images from the training set to create the hold-out set.  We train models for 100 epochs and finally pick the one which performs the best on the held out set. Finally, we see how various models perform against the baseline of 52.79\% \cite{emotiw5} on the validation set, provided by the challenge organizers. The baseline was obtained by training a Support Vector Regression model with a non-linear Chi-square kernel on CENsus TRansform hISTogram (CENTRIST) descriptor \cite{centrist}.

\begin{table}[!htbp]
\centering
\begin{tabular}{l|cccc}
\textbf{Emotion} & \textbf{Positive} & \textbf{Neutral} &\textbf{Negative} &\textbf{Total} \\ \hline \hline
\textbf{Train}       & 1272                                  & 1199                                 & 1159                                  & 3630                               \\ \hline
\textbf{Validation}  & 773                                   & 728                                  & 564                                   & 2065                               \\ \hline
\end{tabular}
\vspace{0.5em}
\caption{5\textsuperscript{th} EmotiW Challenge Dataset}
\label{table:dataset}
\end{table}
\vspace{-2em}

\begin{figure}
  \includegraphics[width=9cm, height = 15cm]{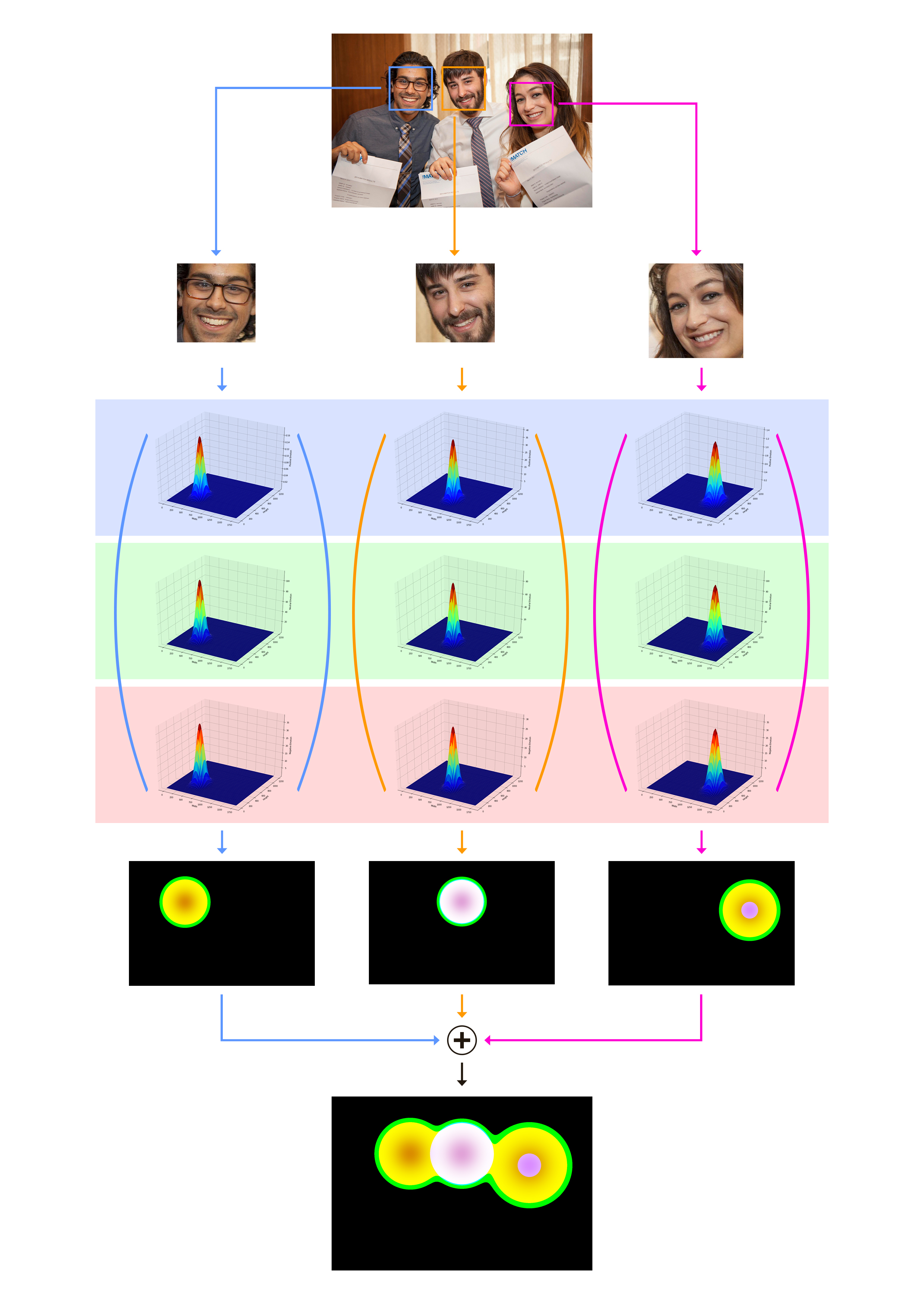}
  \caption{Image Preprocessing Pipeline. From the image, faces are extracted (second row). For every face, all the three categories of emotion are estimated. Gaussain kernels are centered on each face (third, fourth and fifth rows). The kernels for the three categories are stacked to form an RGB image (sixth row) with the distribution for Negative emotion forming the red channel (fifth row) and the distributions for Neutral and Positive emotions forming the green and the blue channels respectively (fourth and third rows). The middle image (sixth row) has a different color than the other two, since the intensities of emotions predicted for the middle face differ from the other two. Note, that the third heatmap (rightmost) in the sixth row is slightly bigger than the other two, since the face in the image is larger. Finally, the RGB images obtained for each face are added to get a combined heatmap (last row).}
  \label{fig:heatmap-generation}
\end{figure}

\section{Pipeline}

\subsection{Face Detection and Cropping}
In order to predict the emotion of an image, we need to detect the emotion of each face in the image. The emotions of individual faces in the image would aid us in predicting the final emotion of the image. We used a popular C++ toolkit, Dlib \cite{dlib09} for this purpose. Dlib toolkit contains pre-trained models for face detection in an image. It provides the coordinates of a rectangular frame that fits the face. The frame coordinates are then used to crop faces from images. An example of face detection in an image via Dlib has been shown in Fig ~\ref{fig:heatmap-generation}

\subsection{Emotion Detection For Individual Faces}
The second step towards solving the problem is to detect emotion of every face detected in section 3.1 above. Detection of emotion of a face is an open problem. To get the predictions for every face, we use a pre-trained model by Levi and Hassner \cite{LH_ICMI15_age}. The advantage of using this model is that it is trained on \textbf{Faces In The Wild} dataset \cite{huang2007labeled} so it incorporates faces in different angles, tones, lighting conditions, partially occluded faces etc. 
It is an ensemble of five models. The models are VGG \cite{vgg16} networks which have been trained on Local Binary Patterns \cite{lbp01} \cite{lbp02}. For a given face, the five models give scores for the seven standard emotions, Anger, Disgust, Fear, Happy, Neutral, Sad and Surprise. The resulting scores for the seven emotions are obtained by averaging the predictions of all models. We then use the values of Happy for Positive class, Neutral for Neutral class and an average of Anger, Disgust, Fear and Sad for the Negative class. The value for Surprise predicted by the model was not used since it was difficult to determine which category it should belong to owing to its ambiguous nature. 

\subsection{Inferring Group Emotions}
Now we arrive at the core problem of this challenge, which is to infer group emotion. We combine the predictions obtained for individual faces, as obtained in section 3.2 to make predictions for the group of people in the image. For this we use the concept of heatmaps which is described in sections 3.3.1 and 3.3.2.

\subsubsection{Interpolation}
For the predicted value of emotion of each face in an image, we construct its corresponding emotion heatmaps. We have used linear distribution and bivariate Gaussian distribution to create the heatmaps in our experiments. Each distribution is created using the value of the emotion predicted in section 3.2 as the value at center. For the bivariate Gaussian, the radius of the face (approximated by calculating half the length of the diagonal of the frame of the face, obtained in section 3.1) becomes the Full Width at Half Maximum (FWHM). Each heatmap represents intensity of emotion for a face distributed in a 2D Euclidean space. The distribution is centered at the face. The size of the 2D space in which the heatmap is generated is same as the size of the image. This technique of representing emotions is useful as it allows us to easily combine the values obtained in section 3.2 for multiple faces, and also perform distance based computations as discussed later in the experiments section.

\subsubsection{Creation of Heatmap}
The distributions for each of the detected face are then combined to form one final image. It is done in three steps. First, we create three matrices (corresponding to each emotion) using the method described in the previous section for each face. The second step is to form a combined spatial distribution for each of the three classes. This is done by stacking the heatmap matrix on top of one another. The three heatmaps form the red, green and blue channels of an RGB image. For our experiments we have arbitrarily chosen the red channel for the negative emotion, green channel for neutral emotion and blue channel for positive emotion. The final step is to add the RGB image tensors obtained for each face to form a combined RGB image. The entire flow of generating a heatmap for an image has been illustrated in Figure ~\ref{fig:heatmap-generation}.

\begin{figure}[!htbp]
  \includegraphics[width=8.5cm, height = 4cm]{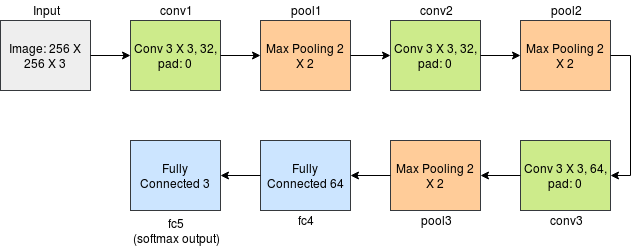}
  \caption{3-ConvNN architecture}
  \label{fig:CNN}
\end{figure}

\subsection{Training Convolutional Neural Networks on Heatmaps}
The size of the images in the EmotiW dataset varies considerably. We first pre-process the images into heatmaps, as described in 3.3, and then resize the images into $256 \times 256 \times 3$ using Python Imaging Library (PIL), since we train Convolutional Neural Networks (CNNs), which require an input of fixed size. We also perform data augmentation/oversampling on the generated heatmaps using Keras'\cite{chollet2015keras} image pre-processing to deal with dataset bias\cite{pixels_to_sentiment}. The augmentations done are random rotations with a range of 40 degrees, random horizontal shifts in the range of 20\% of the total width, random vertical shifts in the range of 20\% of the total height, rescaling the pixel values by a factor of $0.01$, using a shear intensity and zoom intensity of 0.2, random horizontal flips and using the "nearest" fill mode.

Next, we train two different Convolutional Neural Networks, the first of which contains four layers, three of which are convolutional layers and the fourth is a fully connected layer, as depicted in Figure ~\ref{fig:CNN}. We shall refer to it as 3-ConvNN throughout the paper. The first layer of the network takes in images of dimensions $256 \times 256 \times 3$. The first layer consists of 32 filters with a kernel size of $3 \times 3 \times 3$. It is followed by a max pooling layer of size $2 \times 2 \times 3$. The second convolution layer is the same as the first one. The third layer consists of 64 filters with a kernel size of $3 \times 3 \times 3$ followed by a $2 \times 2 \times 3$ max pooling layer. The output of this layer is then flattened and fed to a fully connected layer of size $3$. We also use dropout with a rate of $0.5$. All the layers in the network use Rectifier Linear Unit (ReLU) activation. We use Adam \cite{adam_KingmaB14} for model optimization. The network minimizes categorical cross entropy loss.

\begin{figure}[!htbp]
  \includegraphics[width=9cm, height = 4cm]{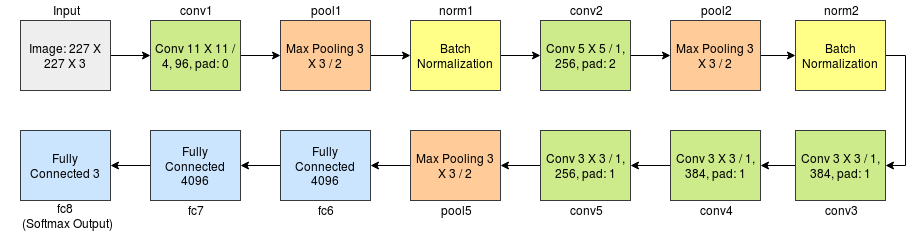}
  \caption{AlexNet Architecture.}
  \label{fig:AlexNet}
\end{figure}
\vspace{-1em}
The second network is AlextNet as described in \cite{NIPS2012_4824}. It takes input with dimensions $227 \times 227 \times 3$. The first convolutional layer contains $96$ filters and a kernel size of $11 \times 11 \times 3$ and a stride of $4$ pixels. It is followed by a max pooling layer of size $3 \times 3 \times 3$ with a stride of $2$ pixels. We then perform batch normalization. Next is a sequence of a zero padding layer of $1$ pixel followed by a convolutional layer of $384$ filters and a kernel size of $3 \times 3 \times 3$ and a stride of 1 pixel. This sequence is repeated again and is followed by a zero padding layer of size $1$ and a convolutional layer of $256$ filters, a kernel size of $3 \times 3 \times 3$ with a stride of 1 pixel. Next, is a max pooling layer of size $3 \times 3 \times 3$ and a stride of $2$ pixels. The output of the max pooling layer is flattened and is fed to a dense layer of $4096$ units. A dropout with a rate of $0.5$ is used next. A dense layer of $4096$ units follows next with which is again followed by a dropout with a rate of $0.5$. 
The final layer is a softmax layer with $3$ units. All the previous laye     rs use ReLU activation. The architecture has been illustrated in the Figure ~\ref{fig:AlexNet}. The model minimizes categorical cross entropy loss. We use Stochatic Gradient Descent (SGD) for model optimization with a learning rate of $0.01$, momentum of $0.9$ and a weight decay of $5 \times 10^{-4}$.

\begin{figure}
  \includegraphics[width=8cm, height = 3cm]{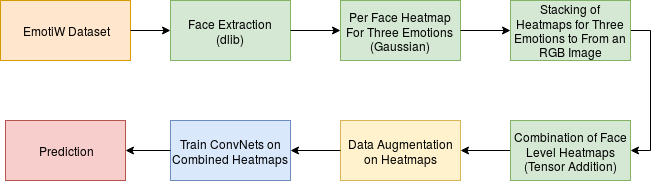}
  \caption{Pipeline of our approach}
  \label{fig:EmotiW-Pipeline}
\end{figure}

The entire approach has been illustrated in Fig ~\ref{fig:EmotiW-Pipeline}

\section{Experimental Analysis}
We now describe a series of experiments performed using raw images as well as the predictions obtained for every face by running the pre-trained model \cite{LH_ICMI15_age}. The experiments in sections 4.1 and 4.2 were done to obtain the group level prediction using predictions for individual faces.
 
\subsection{Using Face Level Predictions Directly}
\subsubsection{Averaging}
	It is one of the simplest approaches that can be used. We take average of 7-dimensional vectors obtained for all the faces in the image. The emotion with the highest mean value is used as the overall prediction. Anger, Disgust, Fear, Sadness and Surprise as categorised as Negative, Neutral and Happy are categorised as Neutral and Positive respectively. However, as discussed in \cite{happiness_intensity}, which is also mentioned in \cite{DBLP_conf_icmi_VonikakisYN016}, averaging is not a reliable measure to predict group level emotion. Upon averaging the face level predictions, we obtained an accuracy of 44.37\% on the training set and an accuracy of 42.38\% on the validation set. 
	
\subsubsection{Random Forest}
	 We next trained a Random Forest classifier (15 estimators) using the averaged 7-dimensional vector obtained for an image. Using Random Forest classifier we achieved an accuracy of 99.08\% on the training set and an accuracy of 48.13\% on the validation set.

\subsection{Training Convolutional Neural Networks on Heatmaps}
	In this subsection we discuss our experiments using the generated heatmaps. Converting images into heatmaps (as discussed in section 3.3) based on face level predictions served two purposes. Firstly, it allowed the ConvNets to focus on isolated face level information by removing the additional information present in the image like the overall scene information. Secondly, it allowed us to perform experiments by considering face sizes and position of faces in space as variables. As we shall see later, taking these factors into account affects the predictions.

\subsubsection{Linear Distribution}
We first use a linear distribution to create heatmaps. The function used is 

\begin{equation}
I(x, y) =  
\begin{cases} 
      \frac{I_0}{d((x, y), (x_0, y_0))} & d \ne 0\\
      I_0 & d = 0
   \end{cases}
\end{equation}

Where $I(x, y)$ is the intensity at the point $(x, y)$ in the image, $I_0$ is the maximum intensity, which is at the center of the face under consideration. This intensity is obtained by using the methodology as described in section 3.2. $d((x, y), (x^{'}, y^{'}))$ is the distance function that calculates the city block distance between the points $(x, y)$ and $(x^{'}, y^{'})$. More specifically, 

\begin{equation}
    d((x, y), (x', y')) = 0.1 \times (|x - x' | + |y - y'|)
\end{equation}

A distance scaling factor of 0.1 is used to make the intensity decrease gradually with distance instead of an abrupt decrease with distance.

Using the above function and the process is as described in section 3.3 and depicted in Fig. ~\ref{fig:heatmap-generation}, we create heatmaps. It is to be noted that, by using a linear distribution, we discard both the size of the faces and their position in space. We train 3-ConvNN on these images.

\subsubsection{Gaussian Distribution}
Until now, the size of the face did not play a role in determining the group level emotion. However, as discussed in \cite{the_more_the_merrier}, large size of faces with smiles play a role in determination of affect of a group. Keeping this in mind, we use a bivariate Gaussian distribution to estimate the intensity of emotion throughout the image. The function which is used to calculate the intensity $I$ at a point $(x, y)$, is

\begin{equation} \label{eqn1}
I(x, y) = I_0 \times exp(\frac{-4 \times ln(2) \times 0.1 \times [(x - x_0) ^ 2 + (y - y_0) ^ 2]}{r})
\end{equation}

Where $I_0$ is the intensity value obtained for a face, $(x_0, y_0)$ is the center of the face in the image and $r$ is the radius of the face which is estimated by half the length of the diagonal of the frame enclosing the face, obtained via face detection. A scaling factor of $0.1$ is again used, to make the intensity decrease gradually in two dimensional Euclidean space.

Using the heatmaps generated by using the above methodology, we 3-ConvNN and AlexNet on the heatmaps. 

\subsubsection{Gaussians Normalized by Distance to Image Center}
We also investigate the effect of position of faces in the image. In order to take position into account we divide the intensity obtained at every point by using a Gaussian distribution by the distance of the center of the face from the center of the image.

More specifically, the intensity at a point $(x_1, y_1)$ is obtained by the following relation:

\begin{equation} \label{eqn2}
I(x, y) = I_0 \times \frac{exp(\frac{-4 \times ln(2) \times 0.1 \times [(x - x_0)^2 + (y - y_0)^2]}{r})}{D((x_f, y_f), (x_c, y_c))}
\end{equation}

Where $D((x_f, y_f), (x_c, y_c))$ is the Euclidean distance between the center of the face under consideration and the center of the image, scaled by a factor of $0.01$. $(x_f, y_f)$ are the coordinates of the center of the face (estimated by the center of the bounding box enclosing the face) in the image and $(x_c, y_c)$ are the coordinates of the center of the image. Thus,

\begin{equation}
    D((x_f, y_f), (x_c, y_c)) = 0.01 \times \sqrt{(x_f - x_c) ^ 2 + (y_f - y_c) ^ 2}
\end{equation}

We then train 3-ConvNN and AlexNet on heatmaps obtained via this method.

\subsection{Training CNNs Directly on Images}
We also trained the two CNNs described before in Section 3.4 on raw images. 3-ConvNN gave a training accuracy of 54.68\% and a validation accuracy of 50.27\%. AlexNet, when directly trained on raw images gave us a training accuracy of 49.57\% and a validation accuracy of 44.98\%. 



\begin{table}[!htbp]
\centering
\small{
\begin{tabular}{l|cc}
\textbf{Model}      & \textbf{\begin{tabular}[c]{@{}c@{}}Training \\ Accuracy\end{tabular}} & \textbf{\begin{tabular}[c]{@{}c@{}}Validation\\ Accuracy\end{tabular}} \\ \hline \hline

\textbf{Baseline}                                                                            & -         & 52.79\% \\ \hline
\textbf{Averaging}                                                                           & 44.37\%   & 42.38\% \\ \hline
\textbf{Random Forest}                                                                       & 99.08\%   & 48.13\% \\ \hline
\textbf{\begin{tabular}[c]{@{}l@{}}Linear Distrbution \\ Heatmaps (3-ConvNN)\end{tabular}}              & 35.59\%   & 38.62\% \\ \hline
\textbf{\begin{tabular}[c]{@{}l@{}}Gaussian Heatmaps \\ (3-ConvNN)\end{tabular}}           & 56.73\%   & 51.49\% \\ \hline
\textbf{\begin{tabular}[c]{@{}l@{}}Gaussian Heatmaps \\  (AlexNet)\end{tabular}} & 57.81\%   & 55.23\% \\ \hline
\textbf{\begin{tabular}[c]{@{}l@{}}Normalized Gaussians\\ (3-ConvNN)\end{tabular}} & 56.89\%   & 54.67\% \\ \hline
\textbf{\begin{tabular}[c]{@{}l@{}}Normalized Gaussians\\ (AlexNet)\end{tabular}}            & 54.51\%   & 52.15\% \\ \hline
\textbf{\begin{tabular}[c]{@{}l@{}}Raw Images\\ (3-ConvNN)\end{tabular}} & 54.68\%   & 50.27\% \\ \hline
\textbf{\begin{tabular}[c]{@{}l@{}}Raw Images\\ (AlexNet)\end{tabular}}            & 49.57\%   & 44.98\% \\ \hline
\end{tabular}
}
\vspace{0.5em}
\caption{Performance of Different Models}
\label{finetuning_1}
\end{table}

\section{Results and Discussion}
Table 2 summarizes the results of various approaches we used to predict group emotion. Averaging of face level predictions discards the information of face sizes and their relative positions in the image. Mean of face level predictions discarding these context features and not working well is also discussed in \cite{happiness_intensity} and \cite{DBLP_conf_icmi_VonikakisYN016}. 

Random Forest classifier (4.1.2) gave an accuracy of 99.08\% on the training set and an accuracy of 48.13\% on the validation set. Though the Random Forest model performed better than the previous approach. We were still behind the baseline accuracy by 4.84\%. The classifier was able to capture complex interplay between various intensities of face level predictions, but this approach, like the former ignores face sizes and relative position of faces. 

Using a linear distribution to create heatmaps (4.2.1) we get an accuracy of 35.59\% on the training set and 38.62\% on the validation set, when the heatmaps are trained on 3-ConvNN. Heatmaps created using a Gaussian distribution (4.2.2) takes the size of the face into consideration. When using that to train 3-ConvNN, we obtain an accuracy of 56.73\% on the training data and 51.49\% on the validation data. Training AlexNet on the same images gave an accuracy of 57.81\% and 55.23\% on training and validation datasets respectively.

When the heatmaps obtained by Gaussians normalized by the distance from the center (4.2.3) are used to train 3-ConvNN, we achieve an accuracy of 56.89\% and 54.67\% respectively on training and validation sets. AlexNet gave an accuracy of 54.51\% and 52.15\% respectively. In this method, both the faces and their positions from the center of the image are considered.

Finally, we see that CNNs trained directly on images do not perform as good as when trained on heatmaps.



\section{Conclusion and Future Work}
This paper presented a pipeline that was used to predict the overall emotion of a group of people in an image for EmotiW 2017 challenge dataset. Ours is a bottom-up approach. We create heatmaps using face-level predictions and train ConvNets on them. We demonstrate that training the CNNs on heatmaps performs better than training them directly on raw images. We also investigate the effects of face size and their position in the image and see that their position does have an impact on the overall affect. Our best model achieves a validation accuracy of  55.23\% which is 2.44\% above the baseline.

Our future work includes different combination methods for heatmaps like the use of manifolds \cite{grassman_manifold}. In addition to that, We also intend to investigate the effect of relative distances between faces in groups.

\section{Acknowledgement}
We thank Balaji Balasubramaniam, Abhishek Sainani and the reviewers for their valuable comments and feedback, which greatly improved our manuscript.

{\small
\bibliographystyle{ieee}
\bibliography{egpaper_for_review.bib}
}

\end{document}